\documentclass{WileyMSP-template}
\usepackage{amsmath}
\usepackage{booktabs}
\usepackage{amssymb}
\usepackage{multirow}
\usepackage{array}
\usepackage{algorithm}
\usepackage{algorithmic}
\usepackage{cleveref}
\usepackage{pifont}
\usepackage{xcolor}
\usepackage{url}
\newcommand{\cmark}{\textcolor{green!80!black}{\ding{51}}}%
\newcommand{\xmark}{\textcolor{red}{\ding{55}}}%
\newcolumntype{C}{>{\centering\arraybackslash}p{1.8cm}}

\begin{document}

\pagestyle{fancy}
\setlength{\headheight}{24pt}

\title{CLIFF: Continual Learning for Incremental Flake Features in 2D Quantum Material Identification}

\maketitle

%
\author{Sankalp Pandey*}
\author{Xuan Bac Nguyen}
\author{Nicholas Borys}
\author{Hugh Churchill}
\author{Khoa Luu*}

\dedication{}

\begin{affiliations}

S. Pandey*, Dr. X.~B. Nguyen, Prof. K. Luu*\\
Quantum AI Lab,\\
University of Arkansas, Fayetteville, AR, USA\\
Email: sankalpp@uark.edu (S.P.); khoaluu@uark.edu (K.L.)

Prof. N. Borys\\
Department of Physics and Astronomy,\\
University of Utah, Salt Lake City, UT, USA\\
Email: nicholas.borys@utah.edu

Prof. H. Churchill\\
Department of Physics,\\
University of Arkansas, Fayetteville, AR, USA\\
Email: hchurch@uark.edu
\medskip\\
\noindent *Corresponding authors.
\end{affiliations}


\begin{abstract}
Identifying flakes of the desired thickness is crucial for selecting candidate materials for quantum devices. Automated layer classification from optical microscopy remains challenging because different thicknesses can produce similar optical signatures, and flake appearance varies across materials. As new materials are introduced, a classifier must learn to recognize their layer classes while retaining its ability to identify flakes from previously learned materials. This paper proposes a new Continual-Learning Framework for Flake Layer Classification (CLIFF). The proposed framework freezes the backbone and base head after training on a reference material. For each new material, it learns a material-specific prompt pool, embedding, and delta head. Together, these components predict the material and thickness from the flake crop alone. Training incorporates memory replay with knowledge distillation. CLIFF achieves higher mean accuracy across learning stages, higher final accuracy, and lower average forgetting than naive fine-tuning and Learning-to-Prompt (L2P) across three training seeds on a grouped four-material benchmark.
\end{abstract}

\section{Introduction}

\begin{figure}
    \centering
    \includegraphics[width=0.9\linewidth]{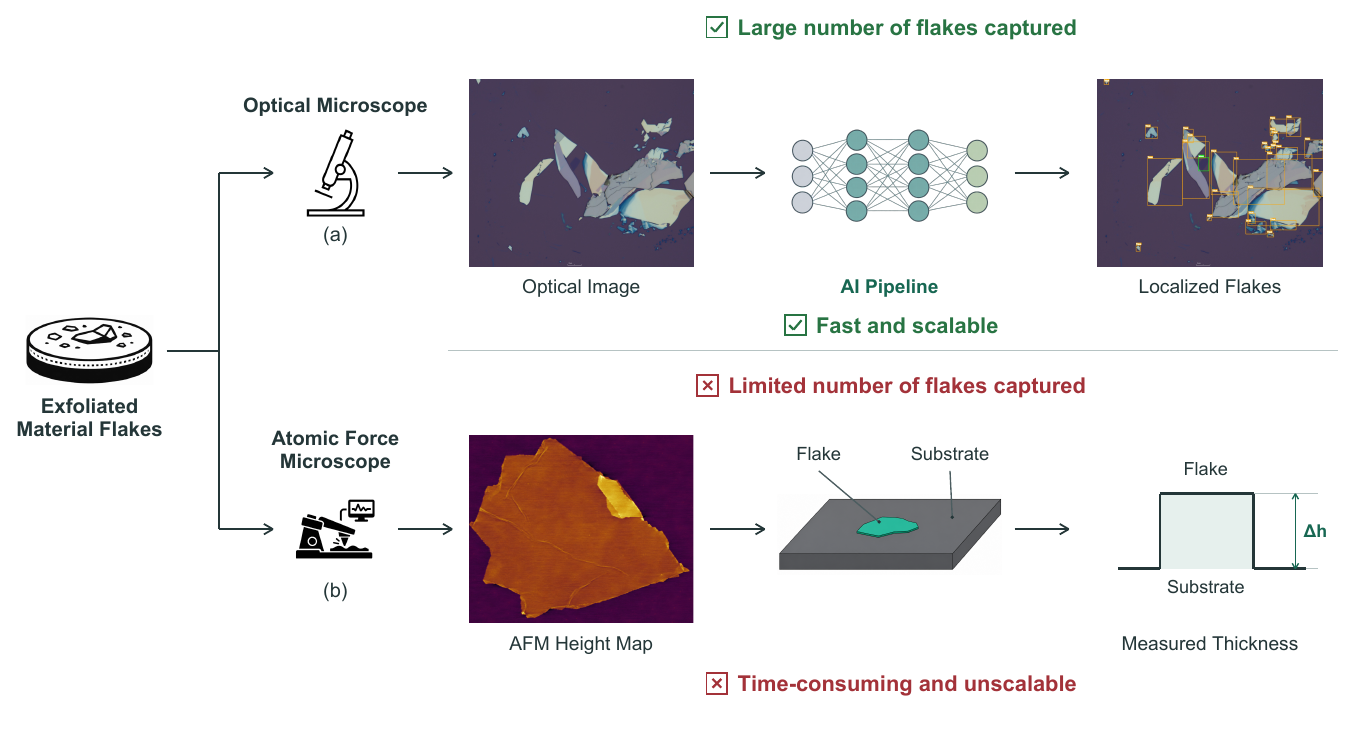}
    \caption{\textbf{Comparison of 2D Material Flake Characterization Methods. }\textbf{(a) }Automated thickness identification using deep learning on optical microscopy images offers a fast, scalable approach capable of processing large numbers of flakes and predicting their layer count. \textbf{(b) }The traditional approach relies on Atomic Force Microscopy (AFM), which is highly accurate but time-consuming and unscalable for high-throughput discovery.}
    \label{fig:measurements}
\end{figure}

Characterizing the layer counts of two-dimensional (2D) material flakes is crucial for the fabrication of van der Waals heterostructures, which facilitate a range of studies and applications \cite{nguyen2025quantumbrainquantuminspiredneuralnetwork, nguyen2025qmoe,nguyen2024qclusformer,holliday2025quadro,nguyen2024quantum,nguyen2024hierarchical,nguyen2025diffusion, holliday2026advanced, holliday2024hybrid,salehi2026unsupervised,nguyen2023quantum,nguyen2026quacod,holliday2025advancedquantumannealingapproach}, especially those in quantum mechanics \cite{james2021recent, lemme20222d,liu2022chemical}. Due to the inherent randomness in the shape and spatial patterns of the flakes, identification and exploration become challenging. Typically, researchers identify flakes through repeated, manual optical microscopy searches, and the samples must be transferred to an Atomic Force Microscope (AFM) for thickness measurements, adding a separate measurement step to heterostructure fabrication. Deep learning approaches aim to overcome manual bottlenecks in classifying flake layers directly from optical images, as highlighted in \Cref{fig:measurements}, but these methods exhibit limited versatility. This work\footnote{The code is available at \url{https://github.com/uark-cviu/quantumflake}.} presents, to the best of our knowledge, the first systematic study of continual learning for 2D flake-thickness classification. The evaluation is set up as a material-incremental, continual learning problem, where new materials arrive sequentially.

\subsection{The Challenges of Automated Flake Identification} The difficulty in automating the identification of 2D materials is estimating the layer count of a flake from optical microscopy images. 
Optical images provide indirect cues to thickness. For example, a monolayer and a bilayer can appear similar under the same imaging conditions. 
Under a regular optical microscope, flakes have faint contrast differences with the substrate (e.g., SiO$_2$/Si), and the contrast between different flake thicknesses can also be subtle and non-linear.
Importantly, the subtle visual characteristics for layer count classification are highly dependent on factors that vary across real-world laboratory settings. First, different 2D materials, e.g., hBN, graphene, MoS$_2$, WTe$_2$, exhibit distinct visual profiles. Second, the appearance of a flake is sensitive to configuration-specific factors, such as illumination conditions and substrate thickness, which may vary between experiments or between datasets. This variability makes it challenging to train a typical deep learning model, motivating the use of a continual learning approach to preserve learned information.

\subsection{Limitations of Prior Work} Traditional and deep learning approaches for flake-layer classification from optical microscopy images have critical limitations. Traditional machine learning methods, such as support vector machines (SVMs) \cite{hearst1998support} and k-nearest neighbors (K-NN) \cite{fix1985discriminatory}, are based on features such as color histograms or texture descriptors, which often fail to represent the subtle and non-linear differences between flakes of different thicknesses. As such, these methods generalize poorly across different imaging setups and materials and are not scalable to large datasets. 
To address some of these challenges, deep learning approaches \cite{han2020deep, masubuchi2020deep} aim to learn discriminative features directly from data, but they still have drawbacks. Obtaining a large labeled dataset is difficult, as it requires tools such as AFM or Raman spectroscopy, which pose a fundamental challenge for deep learning models that require large amounts of labeled data. Models tend to overfit to specific materials or imaging conditions and struggle to adapt to new contexts. Because previous approaches \cite{uslu2024open} were developed and evaluated on their own datasets, their applicability in real-world scenarios is questionable. 
When a pre-trained model is fine-tuned for a new material or experimental setup, it suffers from catastrophic forgetting, in which performance on the original data degrades severely. This inability to learn sequentially without complete retraining makes it impractical for scientific workflows. To our knowledge, continual learning has not been systematically studied for 2D flake thickness classification, and existing models typically degrade when adapting to new materials.

\subsection{Problem Motivation} Addressing these limitations is critical for developing robust automated systems. Although training on a small, fixed set of materials is feasible and can achieve strong performance, it is not a practical strategy for a real-world scientific workflow. In laboratory settings, new materials may be introduced over time, and retraining a large model from scratch with each new dataset can be computationally heavy, especially for larger datasets. Similarly, domain adaptation methods \cite{truong2023fredom, truong2021bimal, nguyen2022self, truong2023comal, 9956335, nguyen2025varphiadaptphysicsinformedadaptationlearning, 9897440, 9710801} are relevant for handling distribution shifts, but they are typically designed for a single source and target domain. This can be insufficient for a growing sequence of new material domains encountered in a laboratory setting. Therefore, flake layer count classification is formulated as a material-incremental continual learning problem where the model must learn new materials while retaining performance on previously seen materials. To address this, this work proposes a continual learning framework that preserves prior knowledge by freezing a backbone and base head trained on a reference material and learning a per-material addition for each new material.


\subsection{Contributions of this Work} 
This work introduces a Continual-Learning Framework for Flake Layer Classification (CLIFF) to jointly identify flake material and thickness from optical microscopy as labeled datasets arrive for new materials.
After training a shared backbone and base classifier on a reference material, CLIFF keeps both frozen and learns a dedicated prompt pool and a residual classifier for each incoming material. At inference, CLIFF scores every seen material and predicts the material-thickness pair directly from the crop. We benchmark sequential adaptation and retention against naive fine-tuning and Learning-to-Prompt (L2P) \cite{wang2022learningpromptcontinuallearning}, scoring all seen materials after each stage. CLIFF achieves higher final macro accuracy and lower average forgetting than both baselines, while naive fine-tuning achieves higher accuracy on newly introduced materials. Across three training seeds, we also examine a second material order and ablate replay, distillation, prompts, and the residual addition in the incremental classifiers.

\section{Related Work}
\subsection{Optical Identification of 2D Material Flakes}

Optical identification of exfoliated flakes relies on how their color and
contrast against the substrate vary with material and thickness. Lin et al.\
used optical color to characterize graphene, MoS$_2$, and their
heterostructures \cite{lin2018intelligent}. Li et al.\ combined
Fresnel-based visibility analysis with RGB clustering and nearest-neighbor
classification \cite{li2019rapid}, while Masubuchi and Machida used
optical and morphological features to identify exfoliated graphene
\cite{masubuchi2019classifying}. Subsequent studies have used unsupervised
segmentation for thickness identification \cite{sterbentz2021universal},
neural networks for monolayer identification \cite{leger2024machine},
and optical images to distinguish one to four layers of WTe$_2$
\cite{he2025machine}. Si et al.\ review the broader use of
machine-learning-assisted microscopy for 2D-material characterization
\cite{si2023correction}. Automated characterization also requires locating suitable flakes within
a microscope field. Han et al.\ used deep learning to identify materials
in optical microscopy images \cite{han2020deep}, and Masubuchi et al.\ used
Mask R-CNN to detect and segment individual flakes
\cite{he2017mask,masubuchi2020deep}. Other approaches use Gaussian mixture
models to classify flakes by optical appearance \cite{uslu2024open} or
self-attention and soft labels to address missing annotations
\cite{nguyen2024two}. More recently, MaskTerial used synthetic images
to pretrain a flake detector \cite{uslu2025maskterial}, Crimmann et al.\
used photoluminescence microscopy for high-throughput identification of
monolayer transition-metal dichalcogenides \cite{crimmann2025high}, and
QuPAINT incorporated optical priors into multimodal instruction tuning
for flake characterization \cite{nguyen2026qupaint}. Other approaches have been proposed for agentic orchestration \cite{pandey2026openqlaw} and domain adaptation as well \cite{nguyen2025varphiadaptphysicsinformedadaptationlearning}. CLIFF takes a flake crop as input and examines how well the model retains its ability to classify the material and thickness of previously encountered flakes as it sequentially learns new materials.

\subsection{Continual Learning}
Continual learning helps prevent catastrophic forgetting in deep learning models. There are several different types of continual learning methods, ranging from scene understanding \cite{truong2023falcon, Truong_2024_CVPR, truong2023fairness, truong2026faidpo} to neuroscience \cite{nguyen2025brainbiasmitigationcontinuallearning, cobraxnguyen}. Rehearsal-based methods store sampled data from previous tasks in buffers \cite{hayes2019memoryefficientexperiencereplay, chaudhry2019efficientlifelonglearningagem, chaudhry2019tinyepisodicmemoriescontinual, 7447103}. As such, they integrate old and new data to preserve knowledge. Knowledge distillation aims to compress knowledge across older and newer tasks \cite{chaudhry2021usinghindsightanchorpast, buzzega2020darkexperiencegeneralcontinual,rebuffi2017icarlincrementalclassifierrepresentation, wu2019largescaleincrementallearning}. Self-supervision is also used \cite{pham2021dualnetcontinuallearningfast, cha2021co2lcontrastivecontinuallearning,9022316, nguyen2024qclusformerquantumtransformerbasedframework}. Some limitations include buffer size, where making it too small limits learning \cite{cha2021co2lcontrastivecontinuallearning} and restricts access to past data \cite{7447103}. Some approaches modify the architecture itself, typically by adding new sets of parameters for new tasks \cite{zhao2021deepbayesianunsupervisedlifelong,li2019learngrowcontinualstructure} or sub-networks specialized for a certain task, which are maintained \cite{wortsman2020supermaskssuperposition, ke2021continuallearningmixedsequence}. Limitations of this approach include the added complexity of the parameters and the model, and the requirement to specify the task type beforehand, which may not be available at test time. In prompt-based learning \cite{wang2022learningpromptcontinuallearning, wang2022dualpromptcomplementarypromptingrehearsalfree}, the backbone is kept frozen, and the model learns a set of task-specific prompt tokens to guide it on new tasks.
\section{The Proposed CLIFF Approach}

\begin{figure}[!t]
    \centering
    \includegraphics[width=0.7\linewidth]{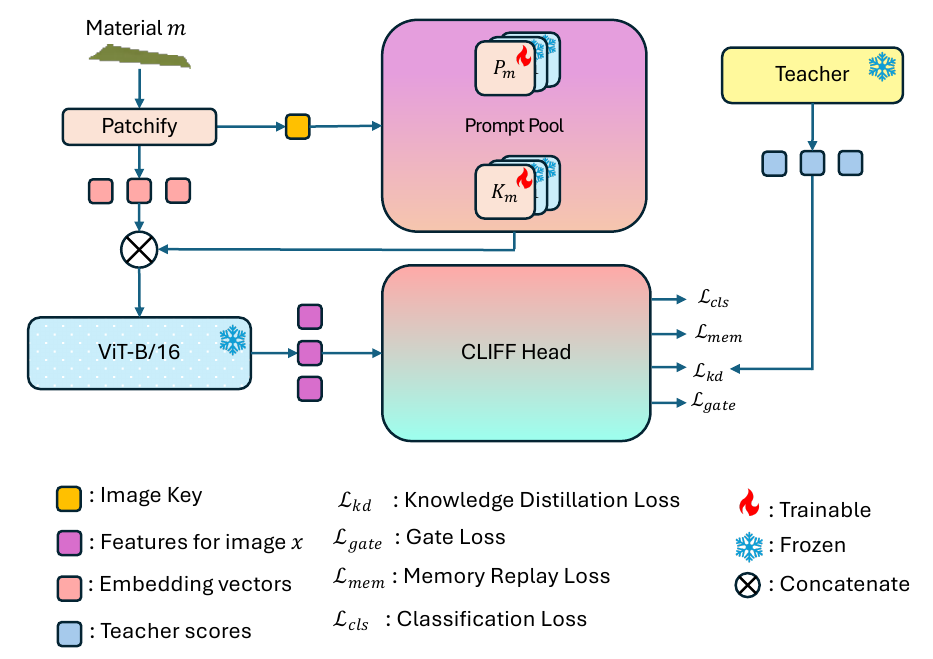} 
    \caption{\textbf{CLIFF Training Overview. }The distillation target uses a reference-stage backbone snapshot and the current prompt and head.}
    \label{fig:clif_overview}
\end{figure}

This work proposes CLIFF, a continual learning framework for 2D flake layer classification. The architecture is visualized in \Cref{fig:clif_overview}. The core idea is to freeze a robust, general-purpose backbone and a base classification head after training on a reference material. For each subsequent material, the framework learns a small set of new components: a dedicated prompt pool, a material embedding, and a delta head that models a material-specific correction. During training on a new material, a prompt pool adapts the frozen backbone's features by prepending learned tokens to the input patch sequence. Training uses replay and a backbone-snapshot distillation term. At image-only inference, CLIFF evaluates every seen material's prompt pool and combines the resulting scores.

\subsection{Base Training on a Reference Material} 
Let $D_1=\{(x_i,y_i)\}_{i=1}^{N_1}$ denote the reference crops and their thickness labels (few-layer, monolayer, or thick). For a crop $x_i$, the Vision Transformer (ViT) \cite{dosovitskiy2021imageworth16x16words} produces the reference-prompted feature $z_i=f_\theta(x_i;P_1)$. The reference stage minimizes as shown in \Cref{eq:lossbase}.
\begin{equation}
\label{eq:lossbase}
 \mathcal{L}_{\mathrm{base}}=\frac{1}{N_1}\sum_{i=1}^{N_1}
 \mathrm{CE}\!\left(g_\phi(z_i)+D_{\mathrm{ref}}([z_i,e_1]),\,y_i\right).
\end{equation}
Here $g_\phi$ is the shared base classifier and $D_{\mathrm{ref}}$ is the reference residual head. During this stage, the backbone, base classifier, and selected prompt tokens receive classification gradients. The reference embedding and residual head remain fixed.




\subsection{Incremental Learning for New Materials}
When a new material $m$ arrives, the approach introduces three new learnable components: a prompt pool $P_m$, a material embedding $e_m$, and a delta head $D_m$.

\subsubsection{Prompting}
Each material has a pool of uniformly initialized prompt tokens and keys. CLIFF retrieves the prompt whose key has the highest cosine similarity to the mean input patch embedding and inserts its tokens before the image patches. The task loss updates the selected tokens. Retrieval uses a hard index, so this loss supplies no gradient to the prompt keys.
The feature output from the prompted backbone is denoted as \Cref{eq:prompt}:
\begin{equation}
\label{eq:prompt}
    z_m = f_{\theta}(x; P_m),
\end{equation} where the output feature $z_m$ is now conditioned on the specific prompt pool $P_m$.

\subsubsection{CLIFF Head}
\begin{figure}[!t]
    \centering
    \includegraphics[width=0.6\linewidth]{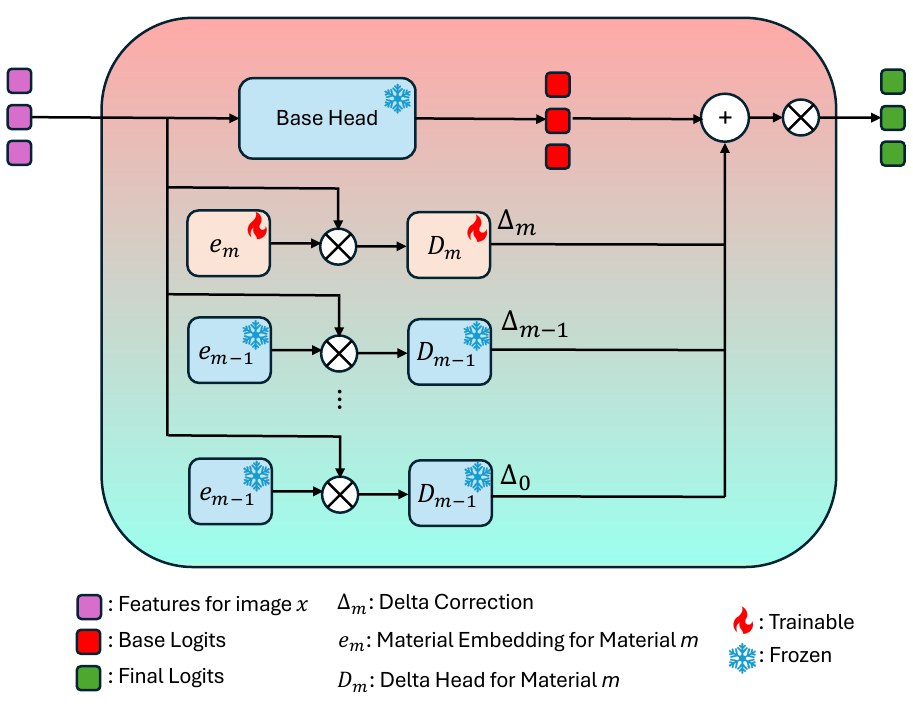} 
    \caption{\textbf{CLIFF Head Breakdown. }CLIFF head during training, when all score blocks receive a common prompted feature. During image-only inference, each block receives the feature from its corresponding material pool.}
    \label{fig:cliff_head_detail}
\end{figure}

 The CLIFF head's architecture is visualized in \Cref{fig:cliff_head_detail}. It processes the prompted features $z_m \in \mathbb{R}^{d}$, where $d$ is the backbone's feature dimension. It maintains an embedding table $E \in \mathbb{R}^{M \times d_e}$ containing a unique embedding vector $e_i$ for each of the $M$ seen materials, where $d_e$ is the embedding dimension. At image-only inference, each material head receives the feature from its own prompt pool. For each material $m$, the Delta Head is implemented as a multilayer perceptron (MLP) consisting of a linear projection to a hidden dimension of 384, a GELU activation, and a Dropout layer ($p=0.2$), followed by a final linear projection. It computes a residual correction $\Delta_m(x) \in \mathbb{R}^{C}$ as described in \Cref{eq:deltaheads}:

 \begin{equation}
 \label{eq:deltaheads} 
 \Delta_m(x) = D_m\!\big(\mathrm{Concat}[z_m,\, e_m]\big). 
 \end{equation} 
 Here, $C=3$ is the number of thickness classes. The final output is a single, large logit vector $L(x)$ created by concatenating the corrected logits for all $M$ materials, as calculated using \Cref{eq:finaloutput}:
 \begin{equation} 
 \label{eq:finaloutput}
 L(x) = \mathrm{Concat}_{i=1}^M \big( g_\phi(z_i) + \Delta_i(x) \big) \in \mathbb{R}^{CM}, \qquad z_i=f_\theta(x;P_i).
 \end{equation}
For each crop, CLIFF retrieves one prompt independently from every seen material pool, runs one transformer pass per pool, and scores each feature with its matching head. The global argmax predicts the material and thickness.

\subsubsection{Optimization with Rehearsal and Distillation}
After the reference stage, the backbone, base head, and gate projection are frozen. Each subsequent stage optimizes the current material's prompt tokens, embedding, and delta head. Both current and replay crops use the current stage's prompt pool during training. All seen heads receive the resulting common feature, and classification cross-entropy is computed over all seen classes.

We add current-stage training crops to replay before training the stage. The buffer holds at most 100 crops per material-thickness class. If the class buffer is full, each additional crop replaces a randomly chosen entry with probability $0.1$. Replay is used after the reference stage. The loss is described as shown in \Cref{eq:loss}:
\begin{equation}
\label{eq:loss}
\mathcal{L}=\mathcal{L}_{\mathrm{cls}}+\lambda_{\mathrm{gate}}\mathcal{L}_{\mathrm{gate}}
+\lambda_{\mathrm{mem}}\mathcal{L}_{\mathrm{mem}}+\lambda_{\mathrm{kd}}\mathcal{L}_{\mathrm{kd}}.
\end{equation}
The classification and replay terms apply cross-entropy to the current and replay batches. The gate term uses cosine-similarity logits between the projected current-crop feature and material embeddings, supervised by the known training material. The reported inference rule uses the score blocks without gate weighting. For the replay distillation target, we use a backbone snapshot taken after the reference stage and apply the current prompt and head under stop-gradient. For target logits $u_T$, student logits $u_S$, and softmax $\sigma$, the term is defined as noted in \Cref{eq:kdloss}:
\begin{equation}
\label{eq:kdloss}
\mathcal{L}_{\mathrm{kd}}=\tau_{\mathrm{kd}}^2\,
\mathrm{KL}\!\left(\sigma(u_T/\tau_{\mathrm{kd}})\,\middle\|\,\sigma(u_S/\tau_{\mathrm{kd}})\right).
\end{equation}

\subsection{Complexity}
\label{sec:complexity}
The total parameter count grows linearly with the number of materials $M$, i.e., $\mathcal{O}(M)$, while the backbone parameters remain constant. For each new material, the added parameters comprise prompt tokens, prompt keys, a material embedding, and a delta head. 
The approximate per-material parameter count is shown in \Cref{eq:per_material_params}:
\begin{equation}
\label{eq:per_material_params}
\underbrace{KLd}_{\text{prompt tokens}}
\;+\;
\underbrace{Kd}_{\text{prompt keys}}
\;+\;
\underbrace{d_e}_{\text{embedding}}
\;+\;
\underbrace{(d{+}d_e)h + hC}_{\text{delta head}},
\end{equation}
where $K$ is the prompt pool size, $L$ is the prompt length, $d$ is the backbone's feature dimension, $d_e$ is the embedding dimension, $h$ is the delta head's hidden dimension, and $C$ is the number of classes. The total parameter count scales as $O(M)$. With $M$ seen materials, the image-only evaluation runs $M$ prompted transformer passes per crop and scores $M$ matching blocks. Transformer compute scales linearly with $M$ at fixed resolution and prompt length. The patch projection is identical across passes and can be shared in an optimized implementation. For rehearsal, a memory buffer that stores $n$ RGB images per class at a resolution of $224\times224$ requires approximately $M \cdot C \cdot n \cdot 224^2 \cdot 3$ bytes of storage.

\section{Experimental Setup}

\subsection{Dataset}
The optical microscopy crops come from the dataset of Masubuchi et al.~\cite{masubuchi2020deep} and cover BN, graphene, MoS$_2$, and WTe$_2$. Each crop carries a material label and one of three thickness labels: few-layer, monolayer, or thick. We grouped crops by material-specific filename-derived source identifiers and kept crops with identical decoded RGB pixels together. This produced 2,070 groups. With split seed 42, we formed five group-disjoint folds stratified by material and thickness. We selected one fold each for validation and test to approximate 20\% of each class in each, and used the other three for training. The resulting partition contains 4,423 training, 1,409 validation, and 1,368 held-out test crops. Every partition contains all twelve material-thickness classes. Physical sample, substrate-region, and acquisition-batch identifiers were unavailable.

\subsection{Evaluation Protocol}
The main comparison introduces BN, graphene, MoS$_2$, and WTe$_2$ in that order. CLIFF, naive fine-tuning, and L2P use the same grouped partition, seeds 42-44, DeiT-Base initialization, image preprocessing, and 15 epochs per material. For each method, we evaluate the checkpoint saved at the end of epoch 15 of each material stage on the grouped test crops. Each prediction uses only the crop and ranges over all material-thickness classes seen by that stage. We use grouped validation for the order comparison and component ablations. We additionally evaluate CLIFF with BN, WTe$_2$, MoS$_2$, and graphene as the arrival order, using the same split, seeds, and epoch budget. The order and ablation experiments span different GPU environments. All variants use the same training seeds and grouped partition. Hyperparameters are fixed across seeds for each configuration.

\subsection{Continual Learning Metrics}
Let $a_{t, i}$ denote the percentage of crops of material $i$ for which the predicted material and thickness are both correct after training stage $t$, with $1\leq i\leq t\leq T$. Following the trajectory metrics in \cite{chen2025sefe}, mean average accuracy (MAA) averages the seen-material macro accuracy across stages, mean first-task accuracy (MFT) averages accuracy on each incoming material immediately after its stage, and mean final accuracy (MFN) is the macro accuracy after the final stage. Backward transfer (BWT) tracks the signed change on earlier materials, and average forgetting (AF) measures their drop from their peak accuracy. 
Each material contributes equally to the macro accuracy, regardless of its number of crops. We compute each metric per seed and report its mean and sample standard deviation.

\subsection{Implementation Details}
The backbone is ViT-B/16 initialized with DeiT-Base weights \cite{deit}. Training crops are augmented with horizontal and vertical flips (probabilities $0.5$ and $0.3$), rotations up to $15^\circ$, and brightness, contrast, and saturation jitter of magnitude $0.2$. Images are resized to $224\times224$ and normalized with ImageNet statistics. Training uses Adam and a current-data batch size of 32. The learning rate is $5\times10^{-6}$ for the backbone during reference training, $5\times10^{-5}$ for the base and delta heads, and $10^{-4}$ for prompts and embeddings. Each delta head has a hidden dimension of 384 and a dropout rate of $0.2$. The material embedding dimension is 128, and each material pool contains 30 prompts of eight tokens each. The replay buffer holds up to 100 crops per material-thickness class, and replay batches contain up to 16 crops. The distillation weight is $1.0$ with a temperature of $3.0$, and the auxiliary gate loss weight is $0.1$.

\subsection{Baselines}
Naive fine-tuning updates the ViT and a linear classifier on the current material, using Adam with learning rates of $5\times10^{-6}$ and $5\times10^{-5}$, respectively, and cross-entropy over all seen classes. The L2P baseline is a PyTorch adaptation of \cite{wang2022learningpromptcontinuallearning}. It freezes the shared DeiT-initialized ViT, retrieves four prompts per image from a shared pool of ten ten-token prompts using frozen CLS features, and classifies the mean of the output prompt tokens. It uses Adam at $0.03$, cosine pull loss, and a current-stage class mask in training. Both baselines train without replay.
\section{Experimental Results and Analysis}

\subsection{Material-Incremental Performance}
\Cref{tab:continual-comparison} compares image-only recognition on held-out crops under a fixed material order. CLIFF achieves the highest mean accuracy over the learning trajectory (MAA) and at the final stage (MFN), and the lowest average forgetting (AF) among the three methods shown. Its final macro accuracy is $36.52$ percentage points higher than naive fine-tuning and $40.60$ points higher than L2P. Naive fine-tuning achieves the highest mean accuracy immediately after learning a material (MFT, $84.80\%$). Its average forgetting is $83.62$ percentage points, compared with $57.39$ for L2P and $24.74$ for CLIFF. CLIFF uses replay and receives more optimizer updates at the shared epoch budget.

\begin{table}[!t]
\centering
\caption{\textbf{Material-Incremental Recognition.} Stage accuracy, MAA, and MFT are percentages. T4 accuracy is the final mean accuracy (MFN). At each stage, accuracy is averaged equally over all materials introduced up to that point. BWT and AF are in percentage points. Values are the mean $\pm$ sample standard deviation across seeds 42-44. Bold marks the best mean in each column.}
\label{tab:continual-comparison}
\label{tab:mat-inc}
\footnotesize
\setlength{\tabcolsep}{2pt}
\begin{tabular*}{\linewidth}{@{\extracolsep{\fill}}l*{8}{r}@{}}
\toprule
 & \multicolumn{4}{c}{Accuracy on All Seen Materials After Each Stage} & \multicolumn{4}{c}{Continual-Learning Metrics} \\
\cmidrule(lr){2-5}\cmidrule(l){6-9}
Method & T1 (BN) & T2 (Graphene) & T3 (MoS$_2$) & \shortstack{T4 (WTe$_2$)\\MFN $\uparrow$} & MAA $\uparrow$ & MFT $\uparrow$ & BWT $\uparrow$ & AF $\downarrow$ \\
\midrule
Naive fine-tuning & $88.65\pm0.61$ & $41.68\pm0.11$ & $31.03\pm1.44$ & $22.08\pm0.31$ & $45.86\pm0.45$ & $\mathbf{84.80\pm0.52}$ & $-83.62\pm1.05$ & $83.62\pm1.05$ \\
L2P & $81.91\pm8.51$ & $40.12\pm9.15$ & $25.47\pm7.77$ & $18.00\pm1.43$ & $41.38\pm5.56$ & $61.05\pm7.85$ & $-57.39\pm11.08$ & $57.39\pm11.08$ \\
\midrule
\textbf{CLIFF (Ours)} & $\mathbf{90.43\pm1.06}$ & $\mathbf{77.32\pm4.95}$ & $\mathbf{68.64\pm3.97}$ & $\mathbf{58.60\pm2.65}$ & $\mathbf{73.75\pm2.55}$ & $77.16\pm1.50$ & $\mathbf{-24.74\pm1.73}$ & $\mathbf{24.74\pm1.73}$ \\
\bottomrule
\end{tabular*}
\end{table}

\subsection{Per-Material Retention}
CLIFF achieves the highest mean accuracy over all seen materials at every stage. After T4, mean accuracy across BN, graphene, and MoS$_2$ is $52.47\%$ for CLIFF, $4.78\%$ for L2P, and $0.00\%$ for naive fine-tuning. On the incoming material WTe$_2$, naive fine-tuning leads with $88.33\%$, followed by CLIFF at $77.00\%$ and L2P at $57.67\%$.

\begin{figure}[!t]
\centering
\includegraphics[width=\linewidth]{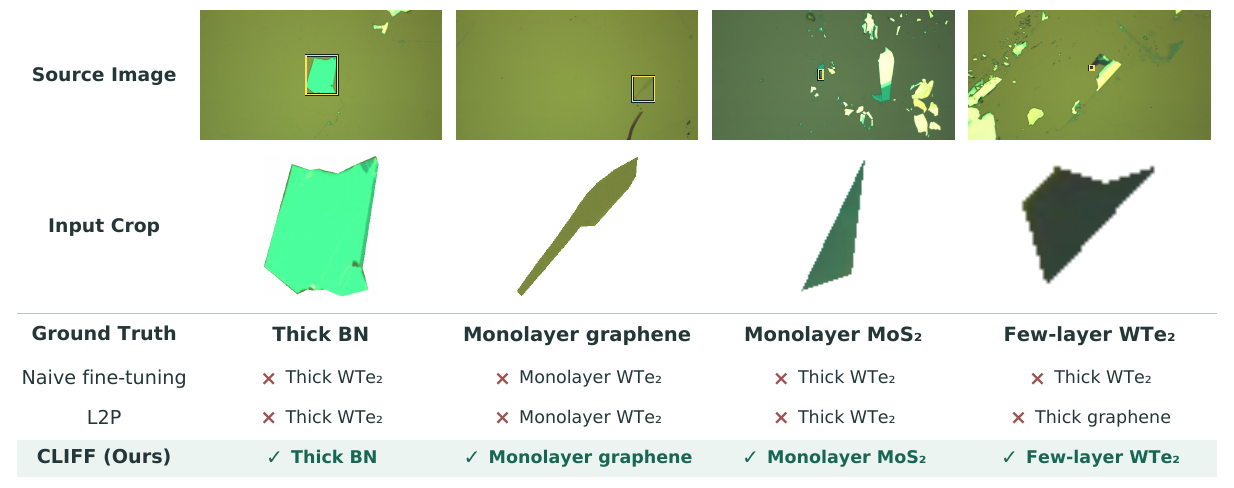}
\caption{\textbf{Qualitative Comparison on Held-Out Crops.} Yellow boxes mark the annotated flake regions in the source images. The row below shows the isolated crops supplied to the models. We selected four examples where CLIFF predicts both labels correctly, and both baselines make an error, using the final-stage checkpoints for seed 42. Checkmarks and crosses indicate correct and incorrect joint predictions.}
\label{fig:qualitative-heldout}
\end{figure}

\Cref{fig:qualitative-heldout} shows selected examples from the final stage.
In the three examples from earlier materials, both baselines predict
WTe$_2$, while CLIFF identifies the correct material and thickness.
On the WTe$_2$ example, CLIFF identifies the few-layer crop and
fine-tuning predicts the thick class.

\subsection{Component Ablations}
\Cref{tab:cliff-ablations} evaluates four changes to CLIFF on grouped validation in the main material order (BN, graphene, MoS$_2$, WTe$_2$). With distillation disabled in both configurations, removing replay reduced mean final accuracy from $62.84\%$ to $18.19\%$ and increased average forgetting from $24.44$ to $76.64$ points. Removing replay also reduces the number of optimizer updates at the fixed epoch budget. Without prompts, mean accuracy immediately after introduction increased from $81.72\%$ to $82.43\%$, while final accuracy fell from $63.49\%$ to $59.40\%$. Removing distillation alone or adding base logits alone changed mean final accuracy by less than one point. With three seeds and variation across runs, these small differences leave the individual benefits uncertain.

\begin{table}[!t]
\centering
\footnotesize
\caption{\textbf{Component Ablations on Grouped Validation.} In Panel A, \protect\cmark{} denotes an enabled component and \protect\xmark{} a disabled component. Scores are the mean $\pm$ sample standard deviation across seeds 42-44. Units and boldface follow \Cref{tab:continual-comparison}. Panel B reports the mean and sample standard deviation of within-seed differences in percentage points (enabled minus disabled). Positive final accuracy and negative forgetting favor the enabled component. The base-logit ablation retains the incremental classifiers and removes the frozen base scores from their outputs.}
\label{tab:cliff-ablations}
\setlength{\tabcolsep}{3pt}
\begin{tabular*}{\linewidth}{@{\extracolsep{\fill}}lccccrrrrr@{}}
\toprule
\multicolumn{10}{@{}l}{\textbf{A. Performance by Configuration}} \\
 & \multicolumn{4}{c}{Components} & \multicolumn{5}{c}{Continual-Learning Metrics} \\
\cmidrule(lr){2-5}\cmidrule(l){6-10}
Configuration & Replay & Distillation & Prompts & \shortstack{Base-logit\\addition} & MAA $\uparrow$ & MFT $\uparrow$ & MFN $\uparrow$ & BWT $\uparrow$ & AF $\downarrow$ \\
\midrule
\textbf{Full CLIFF} & \cmark & \cmark & \cmark & \cmark & $76.31\pm2.41$ & $81.72\pm1.76$ & $\mathbf{63.49\pm3.35}$ & $\mathbf{-24.31\pm2.18}$ & $\mathbf{24.31\pm2.18}$ \\
\midrule
No distillation & \cmark & \xmark & \cmark & \cmark & $\mathbf{77.19\pm1.95}$ & $81.17\pm1.53$ & $62.84\pm5.16$ & $-24.44\pm6.99$ & $24.44\pm6.99$ \\
No replay or distillation & \xmark & \xmark & \cmark & \cmark & $44.33\pm0.63$ & $75.67\pm0.55$ & $18.19\pm0.21$ & $-76.64\pm1.00$ & $76.64\pm1.00$ \\
No prompts & \cmark & \cmark & \xmark & \cmark & $75.94\pm1.85$ & $\mathbf{82.43\pm0.84}$ & $59.40\pm3.45$ & $-30.71\pm3.52$ & $30.71\pm3.52$ \\
No base-logit addition & \cmark & \cmark & \cmark & \xmark & $77.09\pm1.86$ & $81.71\pm1.49$ & $62.51\pm4.44$ & $-25.60\pm6.32$ & $25.60\pm6.32$ \\
\bottomrule
\end{tabular*}
\par\smallskip
\begin{tabular*}{\linewidth}{@{\extracolsep{\fill}}llrr@{}}
\toprule
\multicolumn{4}{@{}l}{\textbf{B. Paired Component Effects}} \\
Component enabled & Comparison condition & $\Delta$MFN $\uparrow$ & $\Delta$AF $\downarrow$ \\
\midrule
Replay & Distillation disabled & $+44.65\pm4.96$ & $-52.20\pm6.03$ \\
Prompts & Other components enabled & $+4.09\pm5.81$ & $-6.40\pm4.78$ \\
Distillation & Other components enabled & $+0.65\pm5.80$ & $-0.13\pm7.31$ \\
Base-logit addition & Incremental heads retained & $+0.98\pm6.71$ & $-1.29\pm7.81$ \\
\bottomrule
\end{tabular*}
\end{table}

\subsection{Replay Buffer Capacity}
\Cref{tab:replay-capacity} varies replay capacity from 20 to 100 crops per material -thickness class with distillation enabled. Mean final accuracy rises from $41.08\%$ to $63.49\%$, while average forgetting falls from $42.89$ to $24.31$ percentage points. The largest gain in final accuracy occurs between 40 and 60 crops per class, from $47.53\%$ to $58.54\%$. The 80-crop buffer achieves the highest MAA ($76.73\%$). Its mean final accuracy and forgetting are close to those at 100 crops per class relative to variation across seeds. Larger buffers can also increase the number of optimizer updates at the fixed epoch budget.

\begin{table}[!t]
\centering
\footnotesize
\caption{\textbf{Replay Buffer Capacity on Grouped Validation.} Capacity is the maximum number of crops stored per material-thickness class. Distillation is enabled at every capacity. Values are the mean $\pm$ sample standard deviation across seeds 42-44. MAA, MFT, and MFN are percentages. BWT and AF are in percentage points. Bold marks the best mean in each column.}
\label{tab:replay-capacity}
\setlength{\tabcolsep}{3pt}
\begin{tabular*}{\linewidth}{@{\extracolsep{\fill}}lrrrrr@{}}
\toprule
Crops per class & MAA $\uparrow$ & MFT $\uparrow$ & MFN $\uparrow$ & BWT $\uparrow$ & AF $\downarrow$ \\
\midrule
20 & $66.04\pm2.01$ & $73.25\pm1.56$ & $41.08\pm5.64$ & $-42.89\pm9.44$ & $42.89\pm9.44$ \\
40 & $68.84\pm1.69$ & $76.40\pm0.52$ & $47.53\pm3.18$ & $-38.50\pm3.61$ & $38.50\pm3.61$ \\
60 & $75.62\pm2.52$ & $79.00\pm1.53$ & $58.54\pm3.93$ & $-27.28\pm4.38$ & $27.28\pm4.38$ \\
80 & $\mathbf{76.73\pm1.67}$ & $79.94\pm0.82$ & $60.72\pm4.30$ & $-25.63\pm6.57$ & $25.63\pm6.57$ \\
\textbf{100 (Full CLIFF)} & $76.31\pm2.41$ & $\mathbf{81.72\pm1.76}$ & $\mathbf{63.49\pm3.35}$ & $\mathbf{-24.31\pm2.18}$ & $\mathbf{24.31\pm2.18}$ \\
\bottomrule
\end{tabular*}
\end{table}

\subsection{Sensitivity to Material Order}
On grouped validation, with the alternative sequence BN, WTe$_2$, MoS$_2$, and graphene, CLIFF achieved $60.56\%$ mean final macro accuracy (sample standard deviation $3.50$ percentage points) and $25.33$ percentage points of average forgetting (sample standard deviation $5.71$ points). On the same validation crops, the main sequence yielded $63.49\%$ and $24.31$ points. These results describe performance across two orders, with BN as the reference material on the same dataset.

\section{Conclusion}
This paper has introduced CLIFF, a continual learning framework for 2D material layer classification that adapts to new materials by learning material-specific prompts and classifiers. The framework incorporates memory rehearsal and knowledge distillation to retain previously acquired knowledge. Experimental results on the grouped four-material benchmark have demonstrated that CLIFF achieves higher mean accuracy across learning stages, higher final accuracy, and lower average forgetting than naive fine-tuning and Learning-to-Prompt (L2P). In practice, CLIFF could help researchers prioritize promising flakes for expert verification as new materials are incorporated into microscopy workflows.

\subsection{Limitations}
Although CLIFF significantly reduces catastrophic forgetting and improves the applicability of continual learning for real-world materials science, a concern with this architecture is scalability. CLIFF adds new parameters for each new material and computes corrections for all seen materials at inference time. While this is feasible within the current scope of this work, linear growth in the number of parameters could lead to heavy computation across hundreds of materials. Additionally, performance depends on the memory buffer, which introduces storage overhead and assumes replayed samples are sufficient to account for feature-space shifts.

\subsection{Broader Impacts} 
CLIFF improves the practicality and usability of automated quantum flake characterization. By enabling deep learning models to adapt to new materials efficiently without forgetting prior knowledge, this approach accelerates discovery in 2D material analysis. It promotes the broader application of robust AI systems in scientific research.


\medskip
\textbf{Acknowledgments} \par 
This work is partly supported by MonArk NSF Quantum Foundry (DMR-1906383), NSF PCL-Test Bed AURORA (2607535) and  NSF Quantum Award (2444042). It acknowledges the Arkansas High-Performance Computing Center for providing GPUs.
\medskip

\bibliographystyle{abbrv}
\bibliography{references}

\end{document}